\documentclass[10pt,twocolumn,letterpaper]{article}

\usepackage{iccv}
\usepackage{times}
\usepackage{graphicx}
\usepackage{amsmath}
\usepackage{amssymb}
\usepackage{microtype}
\usepackage{epsfig}
\usepackage[table,xcdraw]{xcolor}
\usepackage{caption}
\usepackage{float}
\usepackage{placeins}
\usepackage{color, colortbl}
\usepackage{stfloats}
\usepackage{enumitem}
\usepackage{tabularx}
\usepackage{xstring}
\usepackage{multirow}
\usepackage{xspace}
\usepackage{url}
\usepackage{subcaption}
\usepackage[hang,flushmargin]{footmisc}
\usepackage{algorithm}
\usepackage{algorithmicx}
\usepackage{algpseudocode}
\usepackage{bbm}
\usepackage{booktabs}
\usepackage{amssymb}
\usepackage{bbding}
\usepackage{pifont}
\usepackage{makecell}
\usepackage[accsupp]{axessibility}

\usepackage{CJK}
\usepackage[utf8]{inputenc}

\usepackage[breaklinks=true,bookmarks=false]{hyperref}
\iccvfinalcopy 

\ificcvfinal\fi

\begin{document}
\title{SupFusion: Supervised LiDAR-Camera Fusion for 3D Object Detection}

\author{
Yiran Qin$^{1}$\thanks{Equal contribution.  Work done during an internship at NIO.}~, 
Chaoqun Wang$^{1*}$, 
Zijian Kang$^{2}$, 
Ningning Ma$^{2}$, 
Zhen Li$^{3}$, 
Ruimao Zhang$^{1}$\thanks{Corresponding author.}\\
$^{1}$School of Data Science, Shenzhen Research Institute of Big Data, \\The Chinese University of Hong Kong, Shenzhen (CUHK-Shenzhen), China $^{2}$NIO \\
$^{3}$School of Science and Engineering, Future Intelligent Network Research Institute, \\The Chinese University of Hong Kong, Shenzhen (CUHK-Shenzhen), China\\
{\tt\footnotesize \{yiranqin@link, chaoqunwang@link, lizhen@, ruimaozhang@\}cuhk.edu.hk, \{zijiankang, ningningma\}@nio.com}\\
}

\maketitle
\ificcvfinal\thispagestyle{empty}\fi

\begin{abstract}
LiDAR-Camera fusion-based 3D detection is a critical task for automatic driving. In recent years, many LiDAR-Camera fusion approaches sprung up and gained promising performances compared with single-modal detectors, but always lack carefully designed and effective supervision for the fusion process. In this paper, we propose a novel training strategy called SupFusion, which provides an auxiliary feature level supervision for effective LiDAR-Camera fusion and significantly boosts detection performance. 
Our strategy involves a data enhancement method named Polar Sampling, which densifies sparse objects and trains an assistant model to generate high-quality features as the supervision. 
These features are then used to train the LiDAR-Camera fusion model, where the fusion feature is optimized to simulate the generated high-quality features.
Furthermore, we propose a simple yet effective deep fusion module, which contiguously gains superior performance compared with previous fusion methods with SupFusion strategy.
In such a manner, our proposal shares the following advantages. Firstly, SupFusion introduces auxiliary feature-level supervision which could boost LiDAR-Camera detection performance without introducing extra inference costs. 
Secondly, the proposed deep fusion could continuously improve the detector's abilities. 
Our proposed SupFusion and deep fusion module is plug-and-play, we make extensive experiments to demonstrate its effectiveness. Specifically, we gain around $2\%$ 3D mAP improvements on KITTI benchmark based on multiple LiDAR-Camera 3D detectors. Our code is available at \url{https://github.com/IranQin/SupFusion}.

\end{abstract}

\begin{figure}[t]
\begin{center}
\setlength{\abovecaptionskip}{-0.3cm}
   \includegraphics[width=0.99\linewidth]{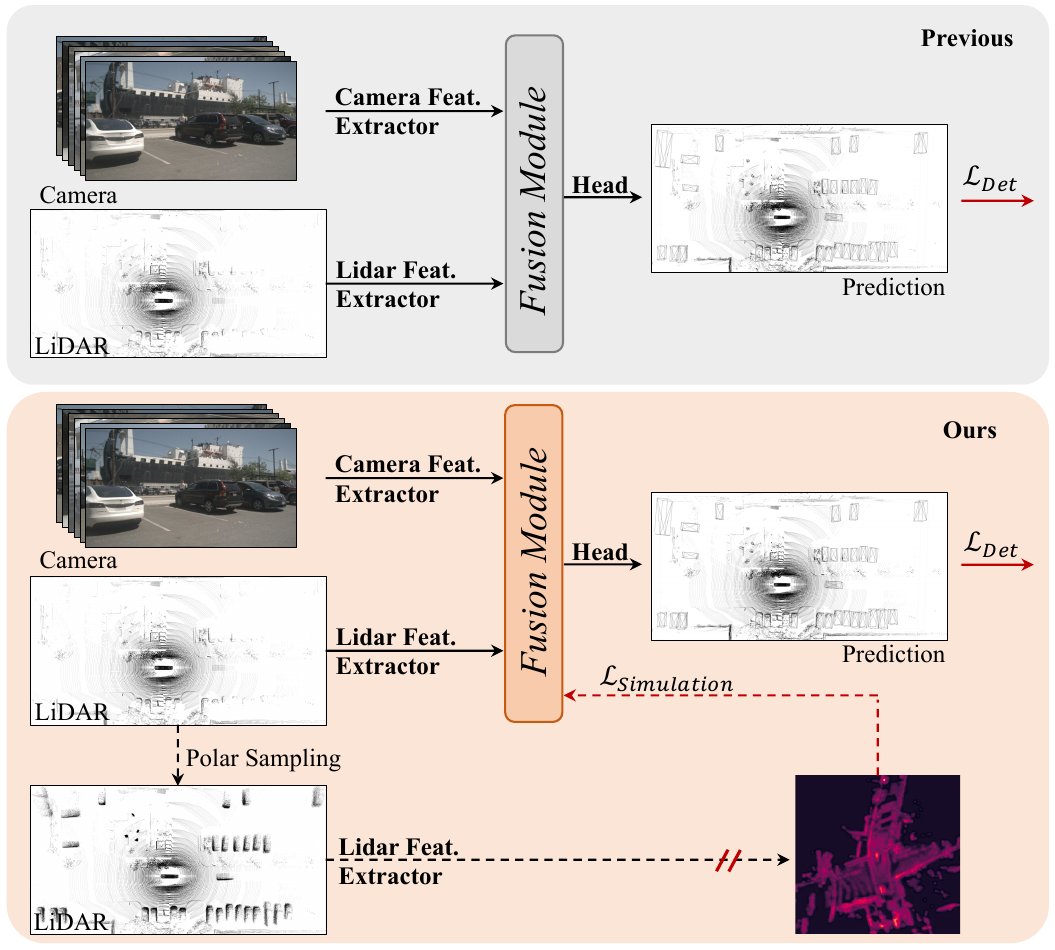}
\end{center}
   \caption{Top: The previous LiDAR-Camera 3D detector, the fusion module is optimized by detection loss. Bottom: Our proposed SupFusion, we introduce auxiliary supervision via high-quality features provided by an assistant model.}
\label{fig:feature}
\end{figure}

\section{Introduction}
LiDAR-Camera fusion-based 3D object detection is a critical and challenging task for autonomous driving and robotics~\cite{sindagi2019mvx,wang2021pointaugmenting}. 
Previous approaches~\cite{li2022voxel,liu2022bevfusion,liang2022bevfusion,li2022uvtr} always project the camera inputs to LiDAR BEV or voxel space via intrinsic and extrinsic to align the LiDAR and camera features.
Then, simple concatenation or summation is employed to obtain the fusion features for final detection. Also, some deep learning-based fusion approaches~\cite{bai2022transfusion,li2022deepfusion} sprung up and gained promising performances. However, the previous fusion approaches always directly optimized the 3D/2D feature extraction as well as the fusion module by the detection loss, which lacks carefully designed and effective supervision at the feature level and limits its performance.

In recent years, distillation manners have shown great improvements in feature-level supervision for 3D detection. For instance, some approaches~\cite{chen2022bevdistill,chong2022monodistill} provide LiDAR features to guide the 2D backbone in estimating the depth information from camera inputs. Also, some approaches~\cite{li2022uvtr} provide LiDAR-Camera fusion features to supervise LiDAR backbone learns global and contextual presentation from LiDAR inputs. 
Introducing feature-level auxiliary supervision via simulating more robust and high-quality features, the detectors could boost marginal improvements. 
Motivated by this,
a natural solution to handle LiDAR-Camera feature fusion is \textit{providing stronger and high-quality features and introducing auxiliary supervision for LiDAR-Camera 3D detection.}

To this end, in this paper we propose a supervised LiDAR-Camera fusion method named SupFusion, to generate high-quality features and provide effective supervision for the fusion as well as feature extraction process, and further boost the LiDAR-Camera fusion-based 3D detection performances.
Specifically, we \textbf{first train an assistant model to provide high-quality features}. To achieve it, different from previous approaches, which leverage larger models or extra data, we proposed a novel data enhancement method named Polar Sampling. The polar sampling could dynamically enhance the object's density from sparse LiDAR data, which is easier to detect and improve features quality, \textit{e.g.}, the feature could conclude accuracy detection results.
Then, we \textbf{simply train the LiDAR-Camera fusion-based detector with introduced auxiliary feature-level supervision}. In this step, we feed the raw LiDAR and camera inputs into 3D/2D backbones and fusion modules to obtain fusion features. In one aspect, the fusion features are fed into the detection head for final prediction, which is decision-level supervision. In another aspect, the auxiliary supervision mimics the fusion feature to high-quality features, which are obtained via the pretrained assistant model and enhanced LiDAR data.
In such a manner, the proposed feature-level supervision could lead the fusion module to generate more robust features, and further boost the detection performance.
To better fuse the LiDAR and camera features, we propose a simple yet effective deep fusion module, which consists of stacked MLP blocks and dynamic fusion blocks. SupFusion could sufficiently excavate the deep fusion module capacity, and continuously improve detection accuracy.

Compared with previous LiDAR-Camera 3D detectors, 
this is a brand new attempt to improve the effectiveness of the LiDAR-Camera fusion by introducing feature-level supervision, which consistently boosts 3D detection accuracy.

The main contributions are summarized as follows.

\begin{itemize}
    \item We propose a novel supervised fusion training strategy named SupFusion, which mainly consists of a high-quality feature generation process and \textbf{firstly propose} auxiliary feature-level supervision loss for robust fusion feature extraction and accuracy 3D detection to our knowledge.

    \item To obtain high-quality features in SupFusion, we propose a data enhancement method named Polar Sampling to densify the sparse objects. Furthermore, we propose an efficient deep fusion module to contiguously boost detection accuracy.

    \item We make extensive experiments based on multiple detectors with different fusion strategies, and gain around $2\%$ mAP improvements on KITTI benchmark. The source code will be released after the blind review.
\end{itemize}
\section{Related Work}

\subsection{3D Object Detection}

Camera-based 3D Object Detection methods perform 3D detection on single- or multi-view images. Many methods are proposed to address monocular 3D detection\cite{lu2021geometry,liu2021autoshape,kumar2021groomed,zhang2021objects,zhou2021monocular,reading2021categorical,wang2021progressive,wang2021depth} since KITTI benchmark\cite{geiger2012we} only contains one from the camera. With the development of autonomous driving datasets which have more cameras covering a wider range of angles(nuScenes\cite{caesar2020nuscenes} and Waymo\cite{sun2020scalability}), more and more methods\cite{wang2022probabilistic,wang2021fcos3d,wang2022detr3d} are developed that take multi-view images as input and get surprising performance compared to monocular methods.

LiDAR-based 3D Object Detection methods perform 3D detection on a point cloud scene. Depending on the three kinds of point cloud data pretreatments,  3D detectors are also classified into three mainstream. Point-based method~\cite{shi2019pointrcnn,yang2019std,zhang2022not,chen2019fast}use row point cloud data, leveraging pointnet\cite{qi2017pointnet} and pointnet++\cite{qi2017pointnet++} as 3D backbone networks to get point-wised features which are sent to point-wise 3D detection head. Pillar-based method \cite{lang2019pointpillars,wang2020pillar,shi2022pillarnet} pretreat point cloud to pillars, gain speed up while still maintaining a good performance. Voxel-based methods\cite{zhou2018voxelnet,yan2018second,deng2021voxel}are widely used, point clouds are voxelized to voxel and sent to a 3D sparse convolutional\cite{graham20183d}  based 3D backbone to get bird's-eye-view features. Some methods\cite{shi2020pv} also try to gain more improvement by using both the voxel feature and the point feature.

\subsection{LiDAR-Camera 3D Object Detection}
LiDAR-camera fusion is a technique that combines data from LiDAR and camera sensors to obtain a more complete understanding of a scene in 3D. Due to the complementary nature of the features produced by these two modalities, researchers have developed methods that can be jointly optimized on both sensors. These methods can be classified into two categories based on their fusion mechanism. The first category is point-level fusion\cite{huang2020epnet,wang2021pointaugmenting,yin2021multimodal}, where the image features are queried via the raw LiDAR points and then concatenated back as additional point features. The second category is feature-level fusion, where the LiDAR points are projected into a specific feature space\cite{yoo20203d,li2022voxel,li2022uvtr} or proposals level feature fusion\cite{chen2017multi,ku2018joint,bai2022transfusion,wu2022sparse}. There are also methods that combine these two fusion manners, such as MVX-Net \cite{sindagi2019mvx}. Recently, attention-based fusion methods have sprung up. TransFusion\cite{bai2022transfusion} uses the bounding box prediction of LiDAR features as a proposal to query the image feature, and then a Transformer-like architecture fuses the information back to LiDAR features. Similarly, DeepFusion\cite{li2022deepfusion} projects LiDAR features on each view image as queries and leverages cross-attention for the two modalities. Bird's-eye-view space feature fusion\cite{liu2022bevfusion,liang2022bevfusion} is proposed recently and gained great performance with different tasks.


\begin{figure*}
\begin{center}
\includegraphics[width=0.99\linewidth]{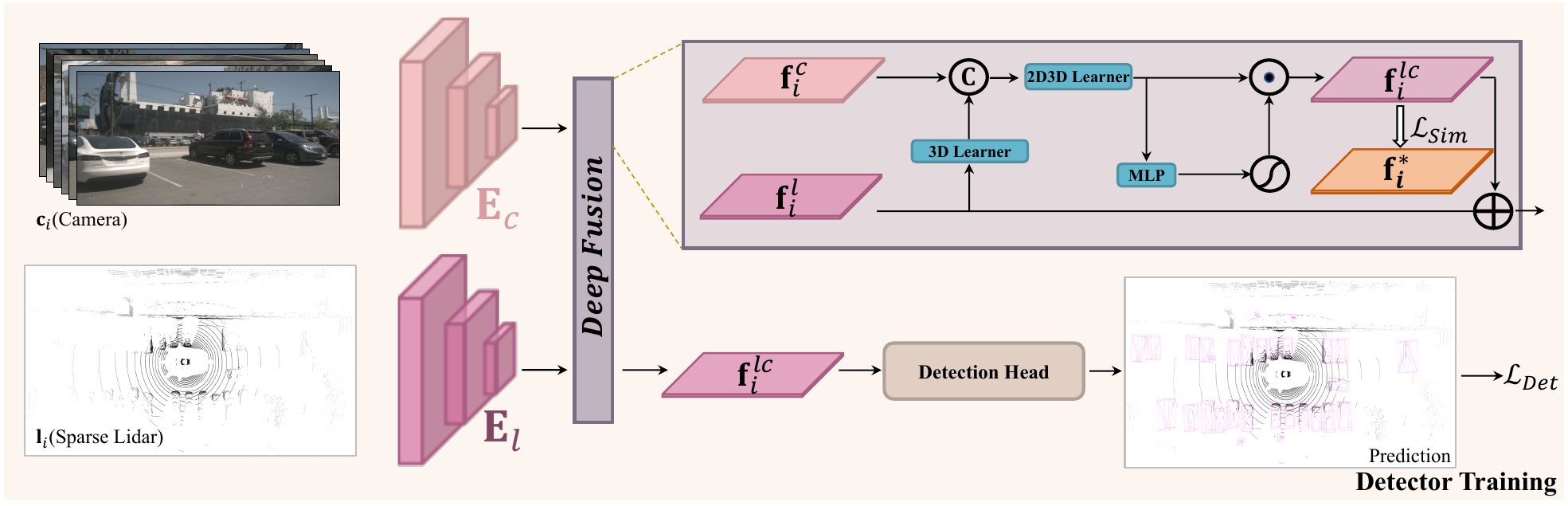}
\end{center}
   \caption{The detector training procedure. For a given LiDAR-Camera sample, we extract the feature presentation via the corresponding 3D/2D encoders, and the proposed deep fusion module is employed to fuse the multi-modal features. Besides the detection loss $\mathcal{L}_{Det}$, we introduce the auxiliary feature-level supervision $\mathcal{L}_{Sim}$, to mimic the fusion feature $\textbf{f}^{lc}_i$ to the high-quality features $\textbf{f}^*_l$, which is generated by the assistant model and enhanced data. Here the $\odot$, $\oplus$, and $\copyright$ indicate the element-wise multiplication, summation, and channel-wise concatenation.}
\label{fig:pipeline}
\end{figure*}

\subsection{Knowledge Distillation for 3D Object Detection}
Object detection tasks differ from classification tasks in that they require consideration of both global and local features. While numerous distillation methods have been proposed to address this challenge, such as those developed by\cite{wang2019distilling,dai2021general,guo2021distilling,zhang2021improve}, many of these methods require the creation of specific distillation regions or objects based on ground-truth data, bounding box predictions, attention, or gradients to mitigate the imbalance between foreground and background. In addition to 2D detection, there has been growing interest in cross-model knowledge distillation methods between RGB and LiDAR in 3D object detection. For example, \cite{guo2021liga,sautier2022image,chong2022monodistill,chen2022bevdistill} have developed methods that transfer knowledge from LiDAR detectors to image detectors. \cite{zheng2022boosting,ju2022paint} has trained a teacher model similar to that proposed by\cite{wang2021pointaugmenting} and developed a multilevel distillation framework to transfer semantic segmentation information to a student model. In this work, we want to further boost LiDAR-Camera detectors with an auxiliary feature level supervision.


\section{Methodology}
In this section, we first overview the overall SupFusion pipeline in Sec.~\ref{sec:Pipeline}, and then detail introduces the main components, including the proposed Polar Sampling in Sec.~\ref{sec:Polar},  and deep fusion module in Sec.~\ref{sec:deepfusion}.

\begin{figure*}
\begin{center}
\includegraphics[width=0.99\linewidth]{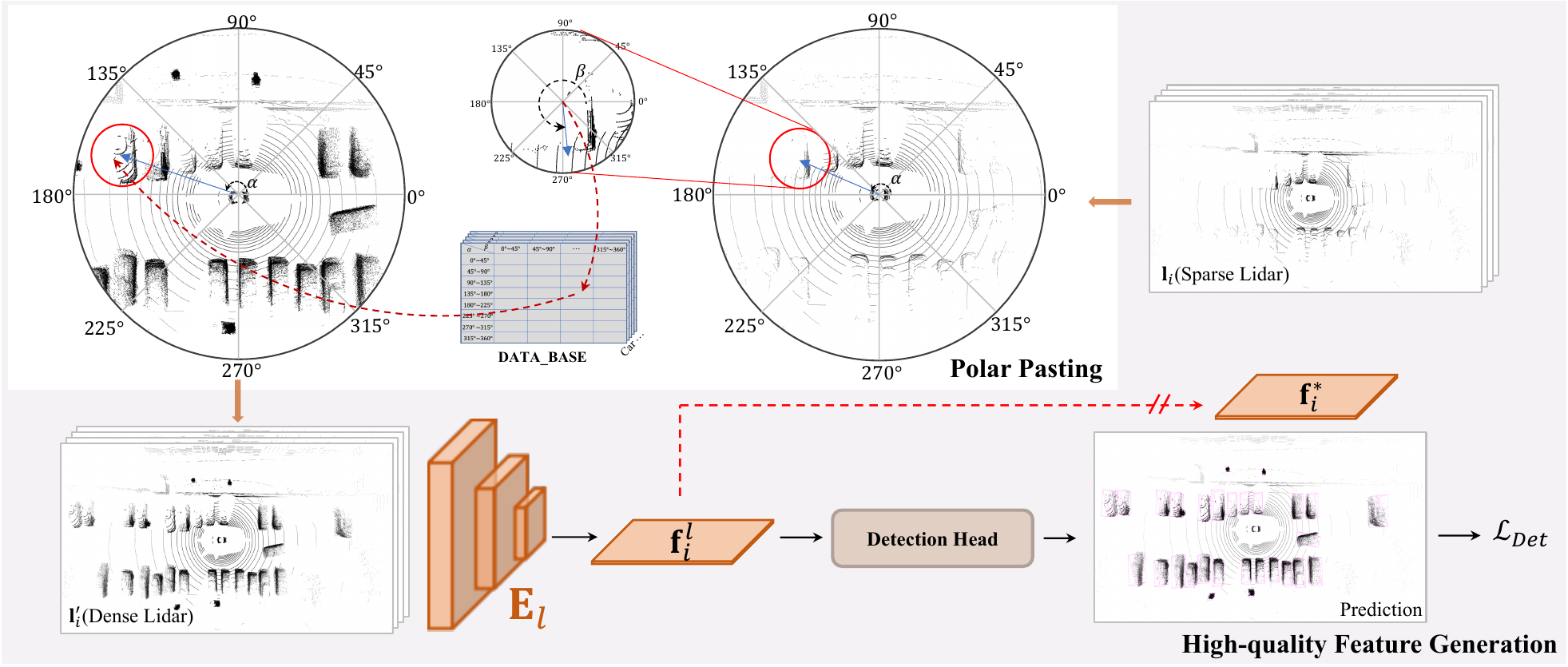}
\end{center}
   \caption{High-quality feature generation process. For any given LiDAR sample, we densify the sparse objects via polar pasting, which calculate the direction and rotation to query dense object from the database, and pasting to add extra points to sparse objects. We first train the assistant model via the enhanced data and feed the enhanced LiDAR data into the assistant model to generate high-quality features $\textbf{f}^*$ after it is convergent. More details could be referred in Sec.~\ref{sec:Pipeline}.}
\label{fig:generation network}
\end{figure*}

\subsection{Overall SupFusion Pipeline}
\label{sec:Pipeline}

For a given LiDAR-Camera fusion-based 3D detector $\mathbf{D}$, which consists of a 3D Encoder $\mathbf{E}_l$, a 2D Encoder $\mathbf{E}_c$, a fusion module $\mathbf{F}$, and a detection head. 
Denotes $\mathcal{D}=\{(\textbf{l}_1, \textbf{c}_1), (\textbf{l}_2, \textbf{c}_2), ..., (\textbf{l}_N, \textbf{c}_N)\}$ and $\mathcal{Y}=\{\textbf{y}_1, \textbf{y}_2, ..., \textbf{y}_N\}$ are the given dataset with $N$ samples, where $(\textbf{l}_i, \textbf{c}_i)$ are the paired LiDAR and camera inputs. The previous approaches feed the paired $(\textbf{l}_i, \textbf{c}_i)$ into 3D/2D encoders $\mathbf{E}_l$ and $\mathbf{E}_c$ to capture corresponding feature presentation $\textbf{f}^l_i$ and $\textbf{f}^c_i$. Then, the paired features presentation is fed into the fusion module to obtain the fusion feature $\textbf{f}^{lc}_i$, which is used for the final prediction. 
To better leverage the LiDAR and camera data, we propose a supervised LiDAR-Camera Fusion training strategy named SupFusion, which provides high-quality features for feature-level supervision. To end this, different from previous approaches, which adopt larger scale models or extra data input, we propose a data enhancement method named polar sampling to densify the sparse objects in LiDAR data. In such a manner, the data is easier and could easily generate high-quality features~(\textit{e.g.}, the feature could conclude accuracy detection results) with lightweight models. 
The proposed polar sampling mainly densifies the sparse object points in LiDAR data.
Specifically, for any sample $(\textbf{l}_i, \textbf{c}_i)$, where $\textbf{l}_i=\{l^1_i, l^2_i,...,l^{n_i}_i,l^s_i\}$ contains $n_i$ objects points sets and one background points set. In the polar sampling, we add extra points to each object point set from raw LiDAR data, to obtain enhanced LiDAR data $\textbf{l}'_i=\{(l^1_i, \hat{l}^1_i), (l^2_i, \hat{l}^2_i),...,(l^{n_i}_i, \hat{l}^{n_i}_i), l^s_i\}$ in polar pasting, where $\hat{l}^k_i$ is the added points set for the $k$-th objects in $i$-th LiDAR sample and generated in polar grouping. We will detail introduce the polar sampling in Sec.~\ref{sec:Polar}. 

\textbf{High-quality Feature Generation.}
To provide feature-level supervision in our SupFusion, we adopt an assistant model to capture high-quality features from the enhanced data as shown in Fig.~\ref{fig:generation network}. Firstly, we train an assistant model to provide high-quality features. Specifically, for any sample in $\mathcal{D}$, we enhance the sparse LiDAR data $\textbf{l}_i$ to obtain enhanced data $\textbf{l}'_i$ via the polar pasting, which densifies the spare objects with added points set generated in polar grouping. Then, after the assistant model convergence~(often with few epochs), we feed the enhanced sample $\textbf{l}'_i$ into the optimized assistant model, to capture high-quality features $\textbf{f}^*_i$, which is used to train the LiDAR-Camera 3D detector. 
To better apply to the given LiDAR-Camera detector and easier implement, we simply adopt the LiDAR branch detector as the assistant model. 

\textbf{Detector Training.}
For any given LiDAR-Camera detector, we train the model with the proposed auxiliary supervision at the feature level. Specifically, given sample $(\textbf{l}_i, \textbf{c}_i) \in \mathcal{D}$, we first feed the LiDAR and camera into the 3D and 2D Encoders $\mathbf{E}_l$ and $\mathbf{E}_s$ to capture the corresponding features $\textbf{f}^l_i$ and $\textbf{f}^c_i$, which are fed into the fusion model to generate fusion features $\textbf{f}^{lc}_i$, and flow into the detection head for final prediction. Furthermore, the proposed auxiliary supervision is employed to mimic the fusion feature with high-quality features, which is generated by the pre-trained assistant model and enhanced LiDAR data. The above process can be formulated as:
\begin{align}
\label{eq:loss}
    \theta^* = \mathop{\arg\min}_{\theta} & \mathbb{E}_{((\textbf{l}_i, \textbf{c}_i), \textbf{y}_i)\sim \mathcal{D}} ~  \mathcal{L}_{det}(\mathbf{D}(\textbf{l}_i, \textbf{c}_i, \theta), \textbf{y}_i) \\
    & + \lambda \mathcal{L}_{Sim}(\textbf{f}^{lc}_i, \textbf{f}^*_i), \\
    s.t.\qquad  \textbf{f}^{lc}_i &= \textbf{F}(\mathbf{E}_l(\textbf{l}_i, \theta_l), \mathbf{E}_c(\textbf{c}_i, \theta_c), \theta_f), \\
    \textbf{f}^*_i &= \mathbf{E}_l(\textbf{l}'_i, \theta^*_l), \\
    \mathcal{L}_{Sim} &= ||\psi(\textbf{f}^{lc}_i) - \textbf{f}^*_i||^2_2
\end{align}
where $\theta^*$ is the optimized parameters of detector $\mathbf{D}$, $\textbf{f}^*_i$ is the high-quality features, which is generated from pre-trained assistant model LiDAR encoder $\mathbf{E}_l$ with parameters $\theta^*_l$, $\psi$ is a simple convolutional layer with kernel size 1. $\lambda$ is the balanced weight and we set it as 1.0.
\subsection{Polar Sampling}
\label{sec:Polar}
To provide high-quality features via the assistant model in our proposed SupFusion, we propose a novel data enhancement method named Polar Sampling, to ease the sparse problem which frequently leads to the detection failers. To end this, we densify the sparse objects in LiDAR data with a similar dense object. Specifically, the polar sampling consists of two parts, including polar grouping and polar pasting. In polar grouping, we mainly build a database to store the dense objects, which is used in polar pasting to densify the sparse objects.

Considering the characteristics of LiDAR sensors, the collected point cloud data naturally exists a particular density distribution, \textit{e.g.}, the objects have more points on the surface towards the LiDAR sensor, while few points on the opposite sides. The density distribution is mainly influenced by the direction and rotation, while the points' density mainly depends on the distances, \textit{e.g.} the objects closer to the LiDAR sensor have denser points. Motivated by this, we aim to densify the sparse objects in longer distances with the dense objects in short distances, according to their direction and rotation to keep the density distributions. Specifically, we build polar coordinate systems for the whole scene as well as the objects based on the center of the scene and specific objects and define the positive direction of the LiDAR sensor as 0 degrees to measure the corresponding directions and rotations. Then, we collect the objects with similar density distribution~(\textit{e.g.} with similar direction and rotation) and generate one dense object for each group in the polar grouping, and used it to densify the sparse objects in polar pasting.

\begin{figure}[t]
\begin{center}
   \includegraphics[width=0.99\linewidth]{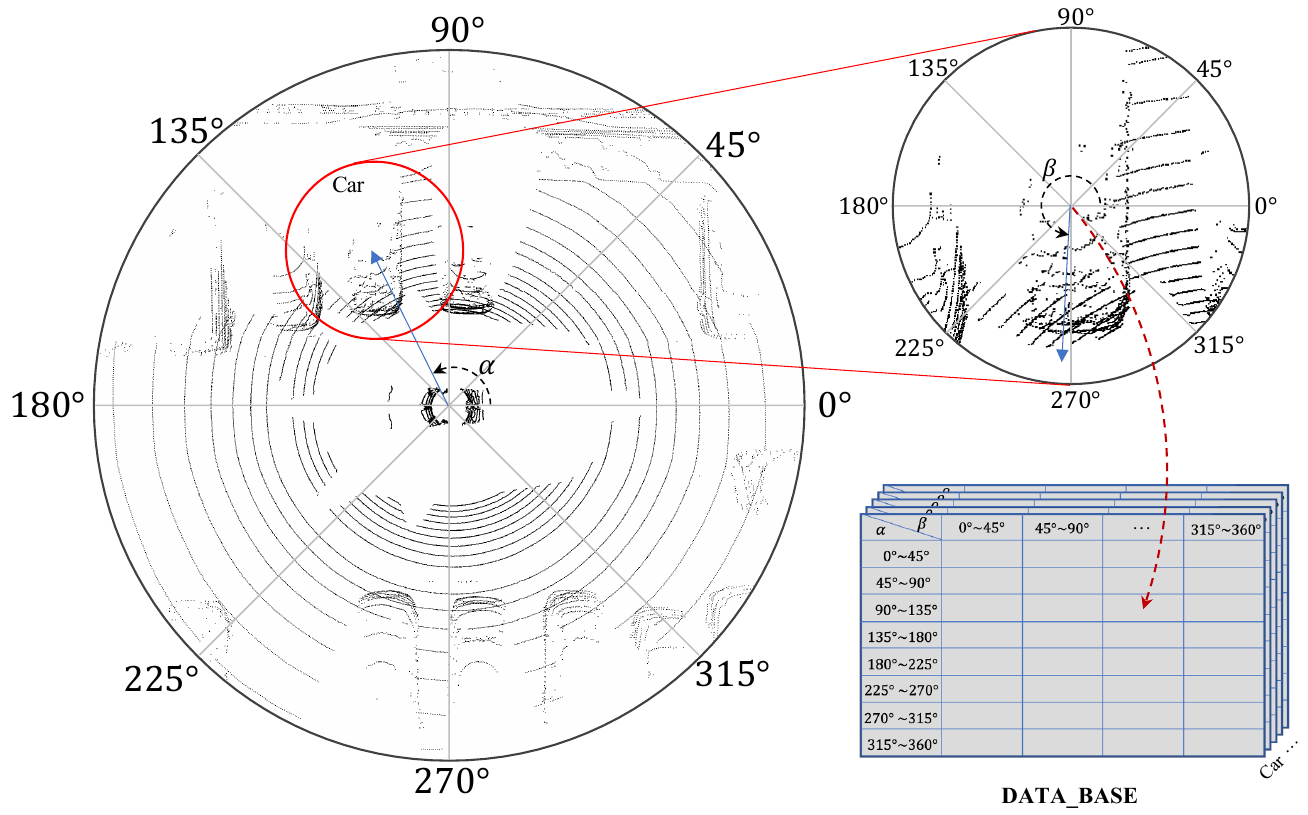}
\end{center}
\vspace{-3mm}
   \caption{Polar grouping mainly builds a database to store the dense objects of each group in polar coordinate systems of direction and rotations for each class. More details could be referred to in Sec.~\ref{sec:Polar}.}
\label{fig:grouping}
\end{figure}

\textbf{Polar Grouping.} 
As shown in Fig.~\ref{fig:grouping}, we build a database $\mathcal{B}$ to store the generated dense object point set $\hat{l}$ according to their directions and rotations in the polar grouping, which is noted as $\alpha$ and $\beta$ in Fig.~\ref{fig:grouping}. 

Firstly, we search the whole dataset and calculated their polar angles for all objects via the locations and provided rotations in the benchmarks.
Secondly, we split the objects into groups according to their polar angles. Specifically, we manually divide the direction and rotation into $N$ groups. For any object point set $l$, we could put it into the corresponding group according to the index:
\begin{align}
\label{eq:align}
    \alpha'_l = \lfloor \frac{\alpha_l \cdot 2\pi}  {N} \rfloor, \quad \beta'_l = \lfloor \frac{\beta_l \cdot 2\pi}  {N} \rfloor,
\end{align}
where $\alpha'_l,\beta'_l\in \{1,2,...,N\}$ are the direction and rotation index of object $l$, and we could put the points set into $\mathcal{B}$ as: $\mathcal{B}^{\alpha'_l, \beta'_l}_C \leftarrow l$, where $C$ is the label of object $l$. Finally, for each object group $\mathcal{B}^{\alpha', \beta'}_C = \{l_1, l_2, ..., l_n\}$ contains $n$ similar objects, we select the densest $k$ objects and mix up to generate one dense and complete object $\hat{l}$, and $\mathcal{B}^{\alpha', \beta'}_C = \{\hat{l}\}$ is used in polar pasting phase.
To keep the density and avoid too much calculation in the model, we randomly select partial points in the dense object. Here we set $k=10$ and select 5000 points for all objects.

\textbf{Polar Pasting.}
As shown in Fig.~\ref{fig:pipeline}, the polar pasting is utilized to enhance the sparse LiDAR data to train the assistant model and generate high-quality features. Concretely, given a LiDAR sample $\textbf{l}_i=\{l^1_i, l^2_i,...,l^{n_i}_i,l^s_i\}$ contains $n_i$ objects,
for any object $l^k_i$, we can calculate the direction and rotation the same as the grouping process and query the dense object $\hat{l}^k_i$ from $\mathcal{B}$ according to the label and indexes, which could be obtained from E.q.~\ref{eq:align} In such a manner, we enhance all the objects in the sample and obtain the enhanced data: $\textbf{l}'_i=\{(l^1_i, \hat{l}^1_i), (l^2_i, \hat{l}^2_i),...,(l^{n_i}_i, \hat{l}^{n_i}_i), l^s_i\}$

\subsection{Deep Fusion}
\label{sec:deepfusion}
To mimic the high-quality features generated from enhanced LiDAR data, the fusion model aims to extract the missing information of the sparse objects from the rich color and contextual features in camera input.
To end this, we propose the deep fusion module, to leverage image features and complete the LiDAR presentation. The proposed deep fusion mainly consists of a 3D learner and a 2D3D learner. The 3D learner is a simple convolutional layer, to transfer the 3D presentation into 2D space. Then, concatenating the 2D features and 3D presentation~(\textit{e.g.} in 2D space), the 2D3D learner is employed to fuse the LiDAR-Camera features. Finally, we weighed the fusion feature via an MLP and an activate function, adding back to the raw LiDAR features as the output of the deep fusion module. The 2D3D learner consists of stacked MLP blocks with depth $K$, learning to leverage camera features to complete the LiDAR representation from the sparse objects, to simulate the high-quality features of dense LiDAR objects.

\begin{table*}[htbp]
\caption{The experiment results~(mAP@R40~$\%$) on KITTI validation set. We list the easy, moderate~(mod.), and hard cases for three classes, and the overall performances. 
Here, L, LC, LC\textbf{*} indicate the results of the corresponding LiDAR detector, LiDAR-Camera fusion detectors, and our proposal. $\Delta$ indicates the improvements. The best results are shown in \textbf{bold} expect L$^\dagger$, which is the assistant model, and test on the enhanced validation set.
MVXNet is re-conduct based on mmdetection3d. PV-RCNN-LC and Voxel-RCNN-LC are re-conduct based on the open-source code of VFF~\cite{li2022voxel}.}
\vspace{-3mm}
\setlength{\tabcolsep}{2mm}{
\renewcommand\arraystretch{1.3}
\begin{tabular}{p{1.0cm}<{\centering}ccccccccccccc}
\bottomrule[1pt]
{\multirow{2}{*}{Detector}} &{\multirow{2}{*}{Modality}}  & \multicolumn{3}{c}{Overall}  & \multicolumn{3}{c}{Pedestrian} & \multicolumn{3}{c}{Cyclist} & \multicolumn{3}{c}{Car}  \\
\arrayrulecolor{gray}\cmidrule(r){3-5} \cmidrule(r){6-8} \cmidrule(r){9-11} \cmidrule(r){12-14}
  {} & {} & easy    & mod.    & hard   & easy   & mod.   & hard  & easy  & mod. & hard & easy   & mod.   & hard  \\
\arrayrulecolor{black}\hline
\multirow{5}{*}{MVXNet} & ~$L^\dagger$& 88.44 & 90.93 & 88.58 & 87.33 & 87.61 & 80.15 & 80.09 & 86.01 & 86.19 & 97.89 & 99.17 & 99.41  \\
{} & $L$ & 73.46 & 58.64 & 55.15&57.54 & 50.10 & 45.56 & 74.63 & 51.81 & 48.92 & 88.21 & 73.99 & 70.98  \\
{} & {$LC$} &74.99 & 63.00 & 59.32 & \textbf{64.46} & 57.76 & 53.05 &72.69 & 55.27 & 51.65 & 87.82 & 75.98 & 73.26  \\
{} & {~~$LC\textbf{*}$} & \cellcolor{cyan!10}\textbf{75.37} & \cellcolor{cyan!10}\textbf{64.73} & \cellcolor{cyan!10}\textbf{60.52} & \cellcolor{cyan!10}63.11 & \cellcolor{cyan!10}\textbf{58.13} & \cellcolor{cyan!10}\textbf{53.73} & \cellcolor{cyan!10}\textbf{74.79} & \cellcolor{cyan!10}\textbf{57.60} & \cellcolor{cyan!10}\textbf{53.82} & \cellcolor{cyan!10}\textbf{88.20} & \cellcolor{cyan!10}\textbf{78.46} & \cellcolor{cyan!10}\textbf{74.01}  \\
\arrayrulecolor{black}
{} & {$\Delta$} & +0.38 & +1.73 & +1.20 & -1.35 & +0.37 & +0.68 & +2.10 & +2.33 & +2.17 & +0.38 & +2.48 & +0.75  \\
\hline
\multirow{5}{*}{\makecell[c]{PV-\\RCNN-LC}} & ~$L^\dagger$ & 94.90 & 95.20 & 95.10&86.86 & 86.78 & 87.07 & 98.90 & 99.33 & 98.62 & 98.95 & 99.50 & 99.62  \\
{} & $L$& 81.37 & 71.55 & 68.08 &67.80 & 59.56 & 54.40 & 85.06 & 70.83 & 67.96 & 91.26 & 84.25 & 81.88  \\
{} & {$LC$}& 84.16 & 74.54 & 70.86& 70.45 & 64.16 & 59.50 & 90.47 & 74.46 & 70.25 & 91.56 & 85.01 & 82.83 \\
{} & {~~$LC\textbf{*}$}& \cellcolor{cyan!10}\textbf{85.77} & \cellcolor{cyan!10}\textbf{76.00} & \cellcolor{cyan!10}\textbf{72.00}& \cellcolor{cyan!10}\textbf{74.29} & \cellcolor{cyan!10}\textbf{66.67} & \cellcolor{cyan!10}\textbf{61.38}& \cellcolor{cyan!10}\textbf{91.00} & \cellcolor{cyan!10}\textbf{76.13} & \cellcolor{cyan!10}\textbf{71.66} & \cellcolor{cyan!10}\textbf{92.02} & \cellcolor{cyan!10}\textbf{85.20} & \cellcolor{cyan!10}\textbf{82.94}  \\
\arrayrulecolor{black}
{} & {$\Delta$} & +1.61 & +1.46 & +1.14 & +3.84 & +2.51 & +1.88 & +0.53 & +1.67 & +1.41 & +0.46 & +0.19 & +0.11  \\
\hline
\multirow{5}{*}{\makecell[c]{Voxel-\\RCNN-LC}} & ~$L^\dagger$ & 99.26 & 99.66 & 99.75& 97.84 & 98.05 & 98.17& 94.36 & 94.60 & 94.85 & 99.90 & 99.90 & 99.91    \\
{} & $L$ & 82.01 & 71.69 & 68.40&67.84 & 60.95 & 56.03 & 86.10 & 71.39 & 67.05 & 92.09 & 82.74 & 82.12   \\
{} & {$LC$}& 84.36 &	74.46 & 70.80 & 71.39 & 64.89 & 60.20 & 89.64 & 72.94 & 69.03 & 92.05 & \textbf{85.55} & \textbf{83.17}  \\
{} & {~~$LC\textbf{*}$}& \cellcolor{cyan!10}\textbf{86.13} & \cellcolor{cyan!10}\textbf{75.90} & \cellcolor{cyan!10}\textbf{72.17 }& \cellcolor{cyan!10}\textbf{73.95} & \cellcolor{cyan!10}\textbf{66.06} & \cellcolor{cyan!10}\textbf{61.83} & \cellcolor{cyan!10}\textbf{92.02} & \cellcolor{cyan!10}\textbf{76.18} & \cellcolor{cyan!10}\textbf{71.83} & \cellcolor{cyan!10}\textbf{92.42} & \cellcolor{cyan!10}85.47 & \cellcolor{cyan!10}82.83 \\ 
{} & {$\Delta$} &  +1.77 & +1.44 & +1.37 & +2.56 & +1.17 & +1.63 & +2.38 & +3.24 & +2.80 & +0.37 & -0.08 & -0.34  \\
\arrayrulecolor{black}\bottomrule[1pt]
\end{tabular}}

\label{tab:overall_kitti}
\end{table*} 

\section{Experiments}
\subsection{Experiments Settings}
\textbf{Datasets and Evaluation Metrics.}
\textbf{KITTI}~\cite{geiger2012we} is a large-scale public benchmark for autonomous driving, including 3712 training, 3769 validation, and 7518 testing samples with LiDAR and camera pair data. All objects in KITTI are divided into 'easy', 'moderate', and 'hard' difficulties according to the object scales and occlusion ratio. We evaluate the detection performance by the mean Average Precision under 40 recall thresholds~(mAP@R40) the same as the official benchmark and evaluation for 3 classes including Pedestrian, Cyclist, and Car. 
\textbf{nuScenes}~\cite{caesar2020nuscenes} is a large-scale autonomous driving dataset for 3D detection and tracking tasks, including 700 training, 150 validation, and 150 testing sequences, with each 20s long and annotated with 3D bounding boxes. For 3D object detection, the official evaluation metrics include the mean Average Precision~(mAP) and nuScenes detection score~(NDS). mAP measures the localization precision using a threshold base on the birds-eye view. NDS is a weighted combination of mAP and regression accuracy of other object attributes, including box size, orientation, and translation.

\textbf{Detectors.} To demonstrate the effectiveness of our proposed method, we apply the SupFusion training strategy and deep fusion module on variance LiDAR-Camera detectors including MVXNet~\cite{sindagi2019mvx}, PV-RCNN-LC~\cite{shi2020pv}, and Voxel-RCNN-LC~\cite{deng2021voxel} on KITTI, and SECOND-LC~\cite{yan2018second} on nuScenes benchmark. 
MVXNet is a voxel-based LiDAR-Camera 3D detector based on VoxelNet~\cite{zhou2018voxelnet}. The camera features project to LiDAR space and concatenates with points features. Voxel-RCNN and PV-RCNN are LiDAR-based detectors, we re-conduct the LiDAR-Camera fusion approach following VFF~\cite{li2022voxel} based on the open-source code, which is a previous state-of-the-art fusion approach based on the two detectors, mainly projects the camera features into voxel presentation and summarized fusion with the voxel features of LiDAR input. SECOND-LC introduces camera inputs based on the LiDAR-based SECOND detector, the same as BEVFusion~\cite{liu2022bevfusion}, we concatenate the camera and LiDAR features in birds-eys-view spaces.

\textbf{Implement Details.} We conduct experiments on MVXNet, PV-RCNN-LC and Voxel-RCNN-LC, SECOND-LC following the open-sourced code and mmdetection3d on 8 Tesla A100 GPUs. All the architecture, configuration, and optimizer follow the settings reported in papers including voxel size, point cloud range, batch size, and learning rate. We modify the fusion method~(\textit{e.g.} concatenation, summarized) by our proposed deep fusion module and applied our SupFusion training strategy. For the deep fusion module, we set depth $K=3$. For the auxiliary supervision, we utilize simple L2 loss and set $\lambda=1.0$. More details about how to apply SupFusion and deep fusion modules on different detectors are demonstrated in the Appendix.

\subsection{Main Results}
\textbf{Overall Performances.}
The 3D mAP@R40 comparison based on three detectors is shown in Tab.~\ref{tab:overall_kitti} for three classes and overall performances for each difficulty split. 
We can clearly observe that the LiDAR-Camera approaches~(LC) are superior to the LiDAR-based detectors~(L) by introducing extra camera inputs. 
By introducing polar sampling, the assistant model~(L$^\dagger$) could gain admirable performances~(\textit{e.g.} over 90\% mAP) on the enhanced validation set. With the auxiliary supervision from high-quality features and our proposed deep fusion module, our proposal could contiguously boost detection accuracy. For instance, compared with the baseline~(LC) model, our proposal could gain +1.54\% and +1.24\% 3D mAP improvements for the moderate and hard objects on average for the three detectors.
Further, we also conduct experiments on nuScenes benchmark based on SECOND-LC as shown in Tab.~\ref{tab:nuscenes}, which gain +2.01\% and +1.38\% improvements for NDS and mAP.
Also, we will prune the SupFusion training strategy and deep fusion module in Sec.~\ref{sec ablation} to detail discuss their effectiveness.

\begin{table}[]
\caption{NDS and mAP(\%) results on nuScenes benchmarks.} 
\vspace{-3mm}
\centering
\begin{tabular}{c<{\centering}|p{1.2cm}<{\centering}p{1.2cm}<{\centering}c<{\centering}c<{\centering} }
\toprule[1pt]
Modality  & 3D &2D & NDS & MAP \\
\hline
  ~$L^\dagger$ &SECOND  & - & 90.38	& 94.44   \\
  $L$ &SECOND & - & 60.52	 & 52.28   \\
  $LC$ &SECOND & Swin-T &62.57	 & 55.26\\
  ~~$LC*$ & \cellcolor{cyan!10}SECOND &\cellcolor{cyan!10}Swin-T &\cellcolor{cyan!10}64.58\tiny(+2.01) & \cellcolor{cyan!10}56.64\tiny(+1.38)\\
\toprule[1pt]
\end{tabular}
\label{tab:nuscenes}
\end{table}

\begin{table}[]
\caption{Results comparison with different fusion methods. Sup. indicates our proposed feature-level supervision. Ped., Cyc., and O.a. are short of Pedestrian, Cyclist, and Overall.}
\vspace{-3mm}
\centering
\begin{tabular}{c|cp{0.7cm}<{\centering}p{0.7cm}<{\centering}p{0.7cm}<{\centering}p{0.7cm}<{\centering}p{0.7cm}<{\centering} }
\toprule[1pt]
Method  & Sup. & Ped. & Cyc. & Car & O.a.  & $\Delta$\\
\hline
\multirow{2}{*}{Sum} &\ding{55}  & 64.89 & 72.94 & 85.55 & 74.46 & -\\
{} & \checkmark{} & 67.03 & 74.61 & 85.11 & 75.59 & +1.13 \\
\arrayrulecolor{black}\hline
\multirow{2}{*}{Concat} &\ding{55}  & 65.90 & 72.80 & 83.37 & 74.02 & -\\
{} & \checkmark{} & 66.23  & 74.95 & 85.42 & 75.52 & +1.50 \\
\arrayrulecolor{black}\hline
\multirow{2}{*}{Ours} &\ding{55}  & 63.18 & 72.84 & 85.58 & 73.86 & -\\
{} & \checkmark{} & \cellcolor{cyan!10}66.06 & \cellcolor{cyan!10}76.18 & \cellcolor{cyan!10}85.47 & \cellcolor{cyan!10}75.90 & \cellcolor{cyan!10}\textbf{+2.04} \\
\arrayrulecolor{black}\hline
\toprule[1pt]
\end{tabular}
\label{tab:fusion_method}
\end{table}

\textbf{Class-Aware Improvements Analysis.} Compared with the baseline model, our SupFusion and deep fusion could boost the detection performances not only for the overall performances but also for each class including Pedestrian, Cyclist, and Car. Comparing the average improvements~(\textit{e.g.} moderate cases) for three classes, we can obtain the following observation: the cyclist gains the largest improvements~(+2.41\%) while pedestrian and car gains +1.35\% and +0.86\% increasements respectively. The reason is obvious: (1) The car is easier to detect and gains the best results compared with pedestrians and cyclists, hence it is more difficult to improve. (2) Cyclist gains more improvements against Pedestrian because pedestrians are non-grid, the generated dense object is not so well compared with Cyclist, hence gains lower performance improvements.

\subsection{Ablation Study}
\label{sec ablation}
To analyze the effectiveness of our proposed SupFusion and deep fusion module, we conduct ablation experiments based on Voxel-RCNN-LC on the KITTI benchmark.

\textbf{Supervision.}
In SupFusion training strategy, we mainly introduce a high-quality guided feature-level supervision and simply employ L2 loss as the auxiliary loss. Here we also adopt L1 loss and the results are shown in Tab.~\ref{tab:sup_loss}, from which we can observe that L1 loss gains comparable but slight improvements~(\textit{e.g.} +1.12\% versus +1.44\%). To mimic the high-quality features by the fusion features, utilizing L1 and L2 to measure the feature distance is acceptable~\cite{he2022masked,xie2022simmim}.

\begin{table}[]
\vspace{-3mm}
\caption{Results comparison with different supervision loss. $\mathcal{L}_{Sim}$ indicates the auxiliary supervision loss in Eq.~\ref{eq:loss}.}
\vspace{-3mm}
\centering
\begin{tabular}{c|cccccc}
\toprule[1pt]
$\mathcal{L}_{Sim}$ & Ped. & Cyc. & Car & O.a. & $\Delta$ \\
\hline
None & 64.89 & 72.94 & 85.55 & 74.46 & -\\
\multirow{1}{*}{L1}  &65.76 & 75.57 &85.42 & 75.58 & +1.12      \\
\multirow{1}{*}{L2} & \cellcolor{cyan!10}66.06 & \cellcolor{cyan!10}76.18 & \cellcolor{cyan!10}85.47 & \cellcolor{cyan!10}75.90 & \cellcolor{cyan!10}+1.44\\
\arrayrulecolor{black}
\toprule[1pt]
\end{tabular}
\label{tab:sup_loss}
\end{table}

\begin{table}[]
\caption{Results with different depth $K$ in deep fusion.}
\vspace{-3mm}
\centering
\begin{tabular}{p{1.05cm}<{\centering}|p{1.05cm}<{\centering}p{1.05cm}<{\centering}p{1.05cm}<{\centering}p{1.05cm}<{\centering}}
\toprule[1pt]
$K$ & 1 & 2 & 3 & 4  \\
\hline
$LC$& 74.23 & 74.00 & 73.86 & 73.22\\
~~$LC$\textbf{*}& 75.64 & 75.75 & \cellcolor{cyan!10}75.90 & 75.88\\
\toprule[1pt]
\end{tabular}
\label{tab:depth_k}
\end{table}

\begin{table}[]
\caption{Results with different $N$ in polar sampling.}
\vspace{-3mm}
\centering
\begin{tabular}{p{1.05cm}<{\centering}|p{1.05cm}<{\centering}p{1.05cm}<{\centering}p{1.05cm}<{\centering}p{1.05cm}<{\centering}}
\toprule[1pt]
$K$ & 1  & 4 & 8 & 16 \\
\hline
~$L^\dagger$ & 99.80 & 99.65 & 99.66 & 99.43 \\
~$LC$\textbf{*} & 75.68 & 75.72 & \cellcolor{cyan!10}75.90 & 75.85\\
\toprule[1pt]
\end{tabular}
\label{tab:split_n}
\vspace{-3mm}
\end{table}

\textbf{Fusion Strategy.}
We conduct experiments with different fusion methods, including the concatenation and summation approaches employed in MVXNet and PV-RCNN-LC/Voxel-RCNN-LC, as well as our proposed deep fusion methods. As shown in Tab.~\ref{tab:fusion_method}, without auxiliary supervision in SupFusion, our proposed deep fusion and concatenation fusion gain comparable but lower performances, demonstrating introducing extra parameters in the fusion model would lead the model difficult to train and hard to optimize. By introducing auxiliary supervision, our proposal could gain the best performance and improvements compared with the baseline model. This demonstrates that the SupFusion could effectively promote the learning process, guiding the 3D/2D backbone and fusion model extracts meaningful robust fusion features from LiDAR and camera inputs. Furthermore, we also manually set variance depth in deep fusion, to rescale the fusion module size. The results are shown in Tab.~\ref{tab:depth_k}, in which we can observe that by increasing the fusion depth, the baseline gets worse while our proposal gains better performances, demonstrating the effectiveness of our proposed SupFusion training strategy and deep fusion module again.

\begin{figure*}[t]
\begin{center}
   \includegraphics[width=0.99\linewidth]{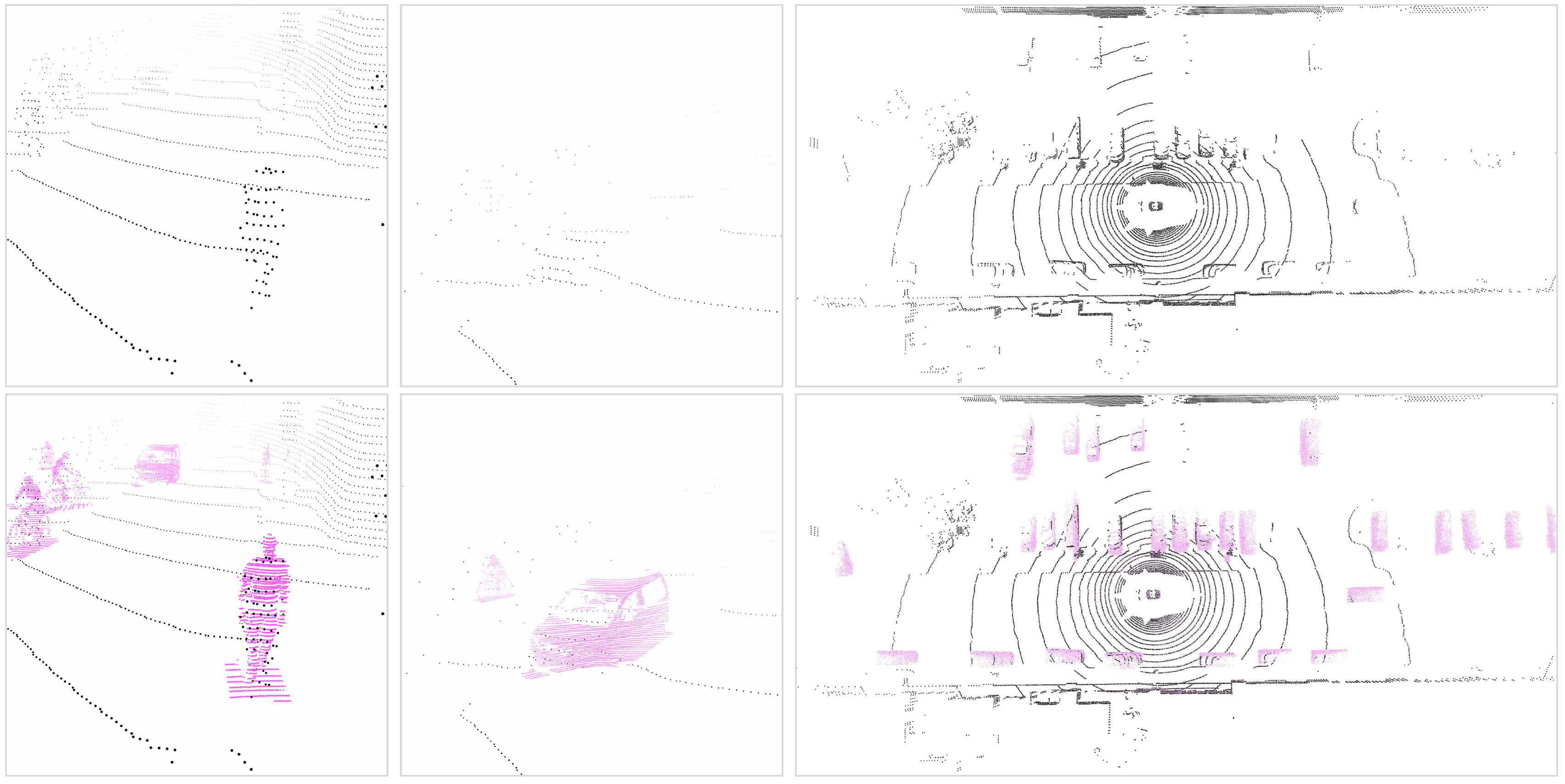}
\end{center}
\vspace{-3mm}
   \caption{Visualization results. The top row shows the raw sparse LiDAR data and the bottom row displays the enhanced data. Particularly, the points in \textcolor[RGB]{255,64,255}{magenta} are the added points and row points are shown in black. We can densify the sparse objects with few points via the dense objects in database while keeping the density distributions.}
\label{fig:vis_polar_sampling}
\end{figure*}

\textbf{Polar Sampling.}
In the proposed polar sampling data enhancement, we split the objects into $N\times N$ groups according to their directions and rotations. To explore the influence of the split number $N$, we manually set $N=\{1,4,8,16\}$ to conduct experiments, and the results are shown as Tab.~\ref{tab:split_n}, from which we can observe that the split number $N$ in polar sampling affect the performances of assistant and our model in an extent. For instance, when setting $N=1$, we only generate one object for each class, the object is complete but destroys the density distribution, which boosts the assistant model performances but decreases the improvements in an extent~(\textit{e.g.} +1.22\% versus +1.44\%), when setting $N$ larger, the objects in each object set may get fewer, because we split all the objects in the dataset into more groups, which may influence the quality of the dense object in the database in the polar grouping.

\textbf{Assitant Model.}
We utilize the enhanced dense LiDAR data to generate high-quality features. By densifying the sparse objects, the assistant model with enhanced data could avoid detection failers easily and provide high-quality features. To better apply to the given LiDAR-Camera detector and easier implement, we simply adopt the LiDAR branch detector as the assistant model. Here, we also conduct experiments to keep the assistant model sharing the same architecture with the given detector, which is a LiDAR-Camera detector. The results are shown as Tab.~\ref{tab:lc_teacher}, from which we can observe that utilizing LiDAR-base and LiDAR-Camera based detectors as the assistant model, the SupFusion gains comparable performances~(\textit{e.g.} LC\textbf{*} versus LC\textbf{**}).

\begin{table}[]
\vspace{-3mm}
\centering
\caption{Results with different teacher models in SupFusion.
LC\textbf{*} and LC{**} indicate utilize LiDAR and LiDAR-Camera based assistant model.}
\begin{tabular}{c|cccccc}
\toprule[1pt]
Model & Ped. & Cyc. & Car & O.a. & $\Delta$ \\
\hline
$LC$ & 64.89 & 72.94 & 85.55 & 74.46 & -\\
$LC$\textbf{*} & \cellcolor{cyan!10}66.06 & \cellcolor{cyan!10}76.18 & \cellcolor{cyan!10}85.47 & \cellcolor{cyan!10}75.90 & \cellcolor{cyan!10}+1.44\\
$LC$\textbf{**} & 66.39  & 75.71  & 85.47  & 75.86 & +1.40\\
\toprule[1pt]
\end{tabular}
\label{tab:lc_teacher}
\end{table}

\subsection{Qualitative Results}
Here we visualize the row sparse LiDAR data and enhanced dense LiDAR in Fig.~\ref{fig:vis_polar_sampling}. In the left 2 columns, we can observe the added points in \textcolor[RGB]{255,64,255}{magenta} could effectively densify the sparse points in black points for all classes including pedestrians, cyclists, and cars. In the right column, we can find that the polar sampling densifies the sparse objects with dense ones while keeping the objects' density distribution. As we could see, enhanced point cloud data have high integrity and obvious class representation which could be utilized to generate high-quality features as the fusion supervision. To better demonstrate the quality of feature-level supervision, we visualize the feature of assistant, baseline, and our model in Appendix.

\subsection{Information leak}There is no information leak if we are not misunderstanding.
In practice, we build the database in Fig.~\ref{fig:grouping} only using the training set in the polar grouping phase, the validation set is not used in polar sampling, as well as the model training process.

\section{Conclusion}
In this paper, we propose a novel training strategy named SupFusion, to introduce auxiliary feature-level supervision for effective LiDAR-Camera fusion, and further boost 3D detection performances. Specifically, we introduce high-quality features to the LiDAR-Camera fusion-based 3D detector, which is used to provide auxiliary supervision for the 3D/2D feature extraction and multi-modal feature fusion. To achieve this, we propose a novel data enhancement method named polar sampling to densify the sparse objects, which is utilized to generate high-quality features via an assistant model. Furthermore, we proposed a simple yet effective deep fusion module, which leverages meaningful information from camera input and contiguously boosts the LiDAR-Camera fusion-based 3D detectors. This is the first approach to introducing feature-level supervision for LiDAR-Camera 3D detection and we conduct extensive experiments to demonstrate its effectiveness.

\section{Acknowledge}
The work is partially supported by the Young Scientists Fund of the National Natural Science Foundation of China under grant No. 62106154, by the Natural Science Foundation of Guangdong Province, China (General Program) under grant No.2022A1515011524, and by Shenzhen Science and Technology Program JCYJ20220818103001002 and by Shenzhen Science and Technology Program ZDSYS20211021111415025.

\clearpage

{\small
\bibliographystyle{ieee_fullname}
\bibliography{egbib}
}

\end{document}